\def\year{2018}\relax
\documentclass[letterpaper]{article} 
\usepackage{aaai18}  
\usepackage{times}  
\usepackage{helvet}  
\usepackage{courier}  
\usepackage{url}  
\usepackage{graphicx}  
\usepackage{amsfonts} 

\usepackage{amssymb}
\usepackage{bbm}
\usepackage{booktabs} 
\usepackage{verbatim} 
\usepackage{amsmath} 
\usepackage[font=small,labelfont=bf]{caption}
\usepackage{stmaryrd}

\usepackage{colortbl}

\newcommand{\ie}{\emph{i.e.,}~}

\begin{document}
%
\title{Exploring Human-like Attention Supervision in Visual Question Answering}
\author{ Tingting Qiao \qquad \qquad Jianfeng Dong \qquad \qquad Duanqing Xu \\
\; Zhejiang University \qquad\qquad \;Zhejiang University \qquad\qquad Zhejiang University\\
{\tt\small qiaott@zju.edu.cn} \quad\quad{\tt\small danieljf24@zju.edu.cn}\qquad\quad{\tt\small xdq@zju.edu.cn}}

\maketitle
\begin{abstract}
Attention mechanisms have been widely applied in the Visual Question Answering (VQA) task,
as they help to focus on the area-of-interest of both visual and textual information. 
To answer the questions correctly, the model needs to selectively target different areas of an image, 
which suggests that an attention-based model may benefit from an explicit attention supervision. 
In this work, we aim to address the problem of adding attention supervision to VQA models. 
Since there is a lack of human attention data, 
we first propose a \emph{Human Attention Network (HAN)} to generate human-like attention maps, 
training on a recently released dataset called Human ATtention Dataset (VQA-HAT). 
Then, we apply the pre-trained HAN on the VQA v2.0 dataset to automatically produce the human-like attention maps for all image-question pairs. 
The generated human-like attention map dataset for the VQA v2.0 dataset is named as \emph{Human-Like ATtention (HLAT)} dataset.
Finally, we apply human-like attention supervision to an attention-based VQA model. 
The experiments show that adding human-like supervision yields a more accurate attention together with a better performance, showing a promising future for human-like attention supervision in VQA.
\end{abstract}

\section{Introduction}

\begin{figure}[tb!]
\centering
\noindent\includegraphics[width=1.0\columnwidth]{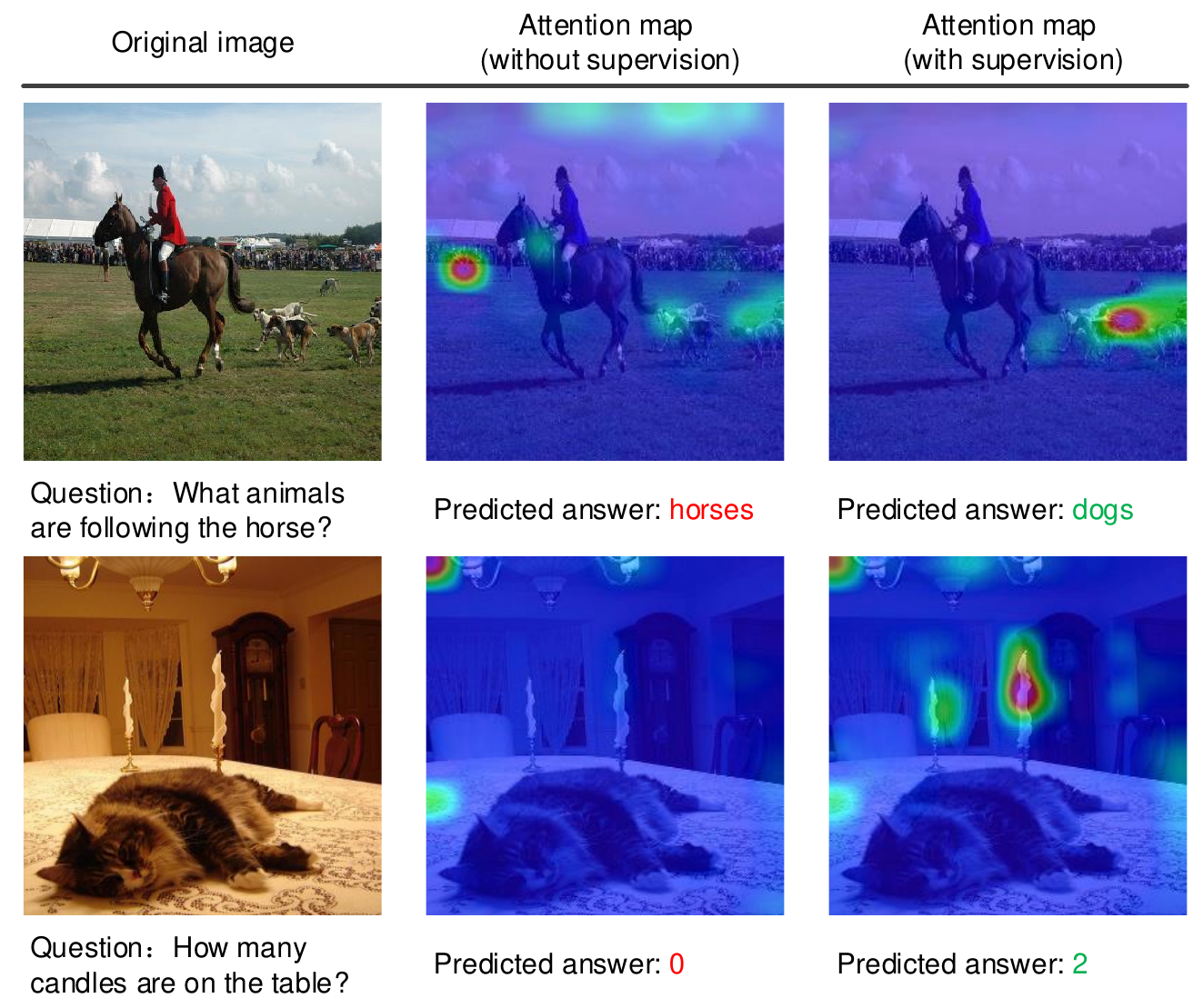}
\protect\caption{Visualization of images, attention maps and predicted answers. As it is possible to see, through explicit attention supervision, the attention maps are more accurate and the predicted answers yield better results. Best viewed in color.}
\label{fig:motivation}
\end{figure}

\noindent
Recently, attention-based models have been proved effective in VQA recently  \cite{shih2016look,yang2016stacked,fukui2016multimodal,lu2016hierarchical,kim2016multimodal}. 
Inspired by human attention mechanisms, these models learn dynamic weights indicating the importance of different regions in an image. 
However, the accuracy of the attention that is implicitly learned during training is not ensured. 
As it can be observed in Fig.\ref{fig:motivation}, 
the model without attention supervision generates inaccurate attention maps, 
which results in the incorrect answers. 
A new VQA-HAT (Human ATtention) dataset has been recently collected by \cite{das2016human} and used to evaluate the attention learned in VQA attention-based models. 
The conclusion of this work was that current attention-based models do not seem to focus on the same area 
as humans do when answering the questions, 
which suggests that there is still room for improving the performance of VQA by adding human attention supervision. 
The public VQA-HAT dataset has good potential to help training. 
However, only around 10$\%$ of the human attention maps for the image-question pairs in the VQA v1.0 dataset \cite{antol2015vqa} are labeled, 
which makes it less effective when used for the purpose of evaluating and improving the quality of the machine-generated attention maps as it is quite limited.
Besides, there are further questions to be explored:
\begin{itemize}
\item Do the regions highlighted by the human attention help VQA attention-based models to yield better answers?
\item How can human attention help to improve the performance of attention-based models in VQA?
\end{itemize}
To tackle these questions, a new and large-scale human attention dataset is first needed. 
Collecting human labeled attention maps is one possible approach, but as mentioned in \cite{das2016human,gan2017vqs}, 
it is a time-consuming and expensive task. 
Since attention-based models in VQA have shown the good performance on VQA tasks,  
we therefore aim to develop a general network to automatically generate attention maps for image-question pairs through training on the existing VQA-HAT. 
We propose a Human Attention Network (HAN) to generate attention maps for image-question pairs. 
Specifically, we propose to use a Gated Recurrent Unit (GRU) \cite{cho2014learning} to encode several attention maps that are learned in the attention-based model during training in order to get a refined attention map. 
The pre-trained HAN is then used to generate attention maps for the image-question pairs in the VQA v2.0 dataset. 
The new attention map dataset for the VQA v2.0 dataset is named as Human-Like ATtention (HLAT) dataset. 

The human-like attention maps are then used as the attention supervision in the VQA attention-based model. 
In order to better verify the effectiveness of the human-like attention supervision, 
we conduct two contrast experiments on the VQA v2.0 dataset. 
To be specific, the attention-based model proposed by \cite{kim2016hadamard} is used as the baseline. 
We then add human-like attention supervision to the baseline model. 
The experiment results show that the human-like attention supervised model outperforms the baseline model by 0.15$\%$, 
demonstrating the effectiveness of human-like attention supervision. 
As it can be seen in Fig.\ref{fig:motivation}, the model with attention supervision generates more accurate attention and gives the better answers. 
To summarize, the main contributions of our work are as follows: 

1. A Human Attention Network (HAN) which can be conveniently used to produce human-like attention maps for image-question pairs is proposed. 

2. The pre-trained HAN is used to produce human-like attention maps for all the image-question pairs in the VQA v2.0 dataset. 
We name the produced attention dataset for the VQA v2.0 dataset as Human-Like ATtention dataset (HLAT). 
The HLAT can be used to explicitly train the attention-based models and also evaluate the quality of the model-generated attention. 
The dataset will be made available to the public. 

3. The overall accuracy of the attention-based model improves 0.15$\%$ on the VQA v2.0 test-dev set by adding human-like attention supervision, showing the effectiveness of explicit human-like attention supervision in VQA. 
Therefore, we get to the conclusion that regions focused by humans indeed help VQA attention-based models to achieve a better performance.
\section{Related Work}
\textbf{VQA attention-based model}
$\;$Traditional models \cite{antol2015vqa,zhou2015simple,noh2016image,andreas2016neural,malinowski2015ask} 
usually present general global features to represent the visual information, 
which may contain a lot of irrelevant and noisy information when predicting the answer. 
Attention mechanisms are therefore proposed to address this problem as
they help models to focus only on the relevant region of the image according to the given question. 
\cite{yang2016stacked} proposed stacked attention networks which employ the attention networks for multi-step reasoning, narrowing down the selection of visual information. 
Dynamic memory networks \cite{kumar2016ask} integrated an attention mechanism with a memory model. 
Multi-model compact bilinear pooling \cite{fukui2016multimodal} combined visual and textual representations to a joint representation. 
Hierarchical co-attention network \cite{lu2016hierarchical} recursively fused the attended question and image features to output the answers. 
Dual attention networks \cite{nam2016dual} employed multiple reasoning steps based on the memory of previous attention. 
\cite{kim2016hadamard} proposed multimodal low-rank bilinear pooling (MLB) which uses element-wise multiplication to reduce complex computations of the original bilinear pooling model. 

One limitation of the mentioned works is that they are unsupervised attention models, whose attention is learned without supervision. 
The attention in these models is always learned implicitly and may end up being simply fortuitous, 
as it can be seen in Fig.\ref{fig:motivation}. 
To address this problem, we explore to add attention supervision to the attention-based model. 
The attention maps in the HLAT are used as attention supervision to train the VQA model on the VQA v2.0 dataset. 

\noindent\textbf{Attention supervision}
$\;$Attention supervision has recently been explored in computer vision. 
\cite{yu2017supervising} leveraged gaze tracking information to provide spatial and temporal attention for video caption generation. 
In image captioning, \cite{liu2017attention} proposed a method to improve the correctness of visual attention using object segments as supervision. 
\cite{gan2017vqs} proposed to use segmentations which are manually labeled by humans to supervise attention generated during training in VQA.
However, using segmentations as attention supervision is potentially inaccurate, especially for some questions that need more reasoning, for example, \emph{are all children paying attention to the class?} Or \emph{Is it a sunny day?} It is ineffective to segment a specific object to answer the questions. Apart from the accurate attention area, the model also needs more global information to analyze. We propose to add human-like attention supervision to models, as human-like attention contains lots of information indicating the important areas that should be focused. 

\noindent\textbf{Human attention dataset}
$\;$Research about human attention and eye tracking has already been carried out \cite{jiang2015salicon,das2016human,gan2017vqs,yu2017supervising}. 
\cite{jiang2015salicon} used mouse-tracking to collect large-scale human attention annotations for MS COCO. 
Recently, \cite{das2016human} published the VQA-HAT (Human ATtention) dataset which contains human attention annotations for around 10$\%$ image-question pairs in the VQA v1.0 dataset. 
These datasets indicate the attentive area of the human when staring at the images and are very useful for evaluation purposes. 
However, as mentioned in \cite{das2016human,gan2017vqs}, 
collecting datasets is always a very time-consuming task and it requires a vast amount of effort. 
In this work, instead of collecting human labeled attention data, the HAN is proposed to generate attention maps automatically, training on the VQA-HAT dataset. 
The pre-trained HAN can be generally used to produce attention maps for image-question pairs.
\begin{figure*}[tb!]
\centering
\noindent\includegraphics[width=2\columnwidth]{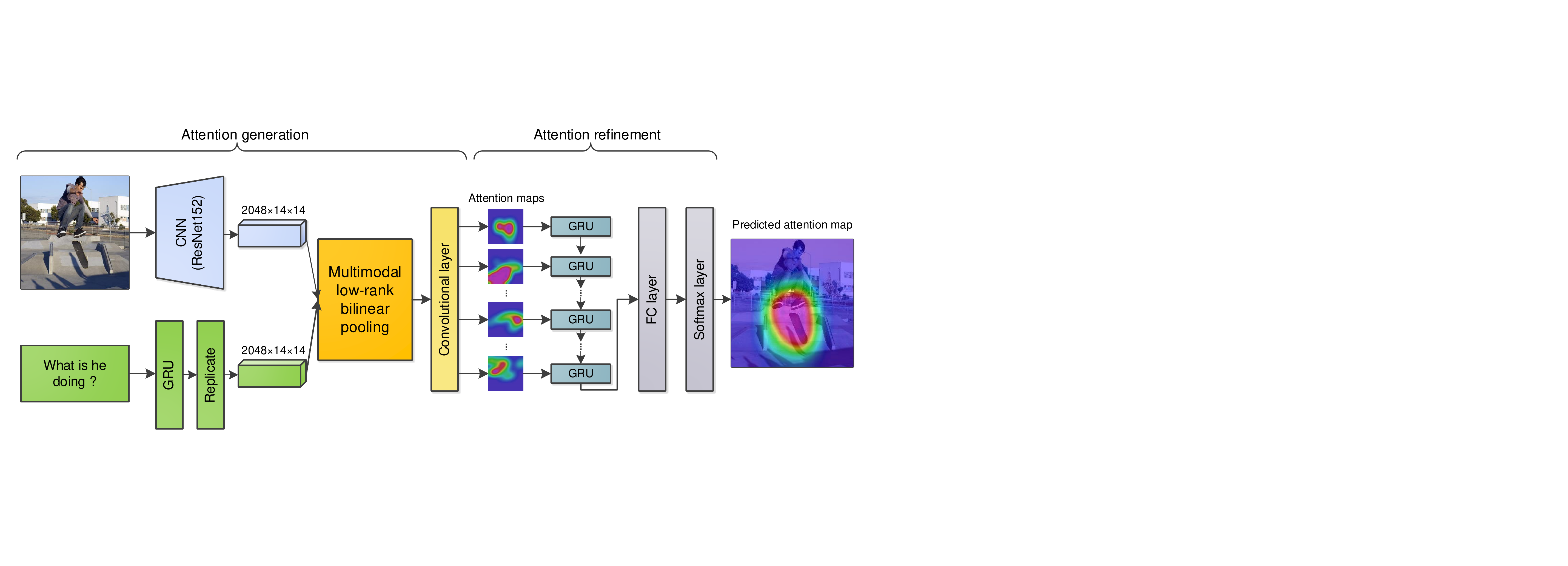}
\caption{Human Attention Network (HAN) for predicting the attention map given an image-question pair.}
\label{fig:han_framework}
\end{figure*} 

\section{Approach}
In this section, details about this work will be provided. 
First, the input representation will be introduced. 
Then, the details of the Human Attention Network (HAN) and the supervised attention model for VQA will be presented.
To help understanding, they are illustrated in different sections.
\subsection{Input representation}
\subsubsection{Image representation}
$\;$Following the common practice in the art \cite{yang2016stacked,kim2016hadamard}, image features are extracted from the ResNet-152 \cite{he2016deep} which is pre-trained on the ImageNet \cite{deng2009imagenet}.
Specifically, the original image $I$ is first scaled to $448\times 448$. Then, the feature from the last convolutional layer of the ResNet-152 is taken as the initial feature map. 
The image $I$ is represented as
\begin{equation}
F=CNN(I),
\end{equation}
where $ F \in \mathbbm{R}^{m \times l^{2}}$ is a feature matrix, $l^{2} = 14 \times 14$ is the number of regions in the image and $m=2048$ is the dimension of the feature vector for each region.
Therefore, the input image $I$ is represented by a set of visual feature vectors over $14 \times 14$ regions of the image.

\subsubsection{Question representation}
$\;$Following \cite{kim2016hadamard}, a GRU is employed to encode the question. 
Specifically, given a question $q = \{ q_1,...,q_{|q|} \}$, where $q_{t}$ is the one-hot encoding representation of the $t$-th word and $|q|$ is the number of words in the question. 
First, the words are embedded to a vector space by a word embedding matrix $W$ which results in $x_t = W q_t$. 
The word embedding vectors are then sequentially fed into a GRU module to get a sequence of hidden state vectors $\{h_1, h_2,..., h_{|q|}\}$, 
\begin{equation}
h_t=GRU(x_{t}, h_{t-1}),t \in \{1,2,...,|q|\}. 
\end{equation}
Finally, the last hidden state vector $h_{|q|}$ of the GRU is taken as the representation of the question, 
\ie $Q=h_{|q|}\in \mathbbm{R}^{n}$, where $n$ is the size of the GRU. 
Note that the word embedding matrix and the GRU are trained end-to-end. 

\subsection{Human Attention Network}
Given an image-question pair $(I,q)$, our proposed Human Attention Network (HAN) is designed to predict the attention map. 

The structure of the HAN is shown in Fig.\ref{fig:han_framework}. It consists of two modules, namely the attention generation module and the attention refinement module. The attention generation module is able to generate several coarse attention maps. 
The attention refinement module aims to refine the coarse attention maps in order to obtain a more precise attention map. 

\subsubsection{Attention generation}
$\;$The attention mechanism used is motivated by \cite{kim2016hadamard}, 
which first fuses the image and the question by a low-rank bilinear pooling and then uses a convolutional layer to generate multiple attention maps over the image. 
Specifically, the feature matrix $F$ of an image $I$ and the feature vector $Q$ of the question $q$ are first projected into a $d$-dimensional embedding space.
Then, they are fused by element-wise multiplication. 
The fused feature matrix of the image and the question is given by 
\begin{equation} \label{eq:att_fusion}
X  = \sigma (U Q\cdot\mathbbm{1}^{T}) \circ \sigma(V F),
\end{equation}
where $\mathbbm{1}\in \mathbbm{R}^{l^{2}}$ denotes a column vector of ones, $U\in \mathbbm{R}^{d\times n}$ and $V\in \mathbbm{R}^{d\times m}$ are respectively trainable affine matrices for the question and for the image. 
The symbol $\circ$ denotes the element-wise multiplication, while $\sigma(\cdot)$ is the hyperbolic tangent function.
Note that the bias terms in the low-rank bilinear pooling are omitted for simplicity. 

Subsequently, a convolutional layer is applied to the fused feature $X$ to generate a sequence of attention maps which is expressed as 
\begin{equation} \label{eq:att_generation}
\alpha_{1}, \alpha_{2},...,\alpha_{G} = C_{con}(X),
\end{equation}
where $C_{con}(\cdot)$ denotes the 1-D convolution operation with $G$ convolutional kernels of size 1. 
In this way, multiple attention maps, also known as multiple glimpses in \cite{kim2016hadamard,fukui2016multimodal}, are generated. 

\subsubsection{Attention refinement}
$\;$The multiple attention maps generated by the attention generation module are coarse, 
however, they contain lots of information. 
In this section, 
the multiple attention maps are fused by means of a GRU, as it has the ability of neglecting irrelevant information and keeping the important information.
Specifically, the sequence of attention maps $\alpha=\{\alpha_{1}, \alpha_{2},...,\alpha_{G}\}$ that is generated by the attention generation module is fed into the GRU sequentially. 
The last hidden state vector of the GRU is taken as the attention vector $h_{GRU}(\alpha) \in \mathbbm{R}^p$, where $p$ is the
size of the GRU. 
The vector $h_{GRU}(\alpha)$ is then driven through a fully connected layer followed by a softmax layer.
More formally, the predicted refined attention map for the given image-question pair $(I,q)$ is calculated as 
\begin{equation}
\alpha_{h}'(I,q)=softmax(W \cdot h_{GRU}(\alpha) + b),  \label{eq:att_gru}
\end{equation}
where $W \in \mathbbm{R}^{l^{2} \times p}$ is an affine transformation matrix, $softmax(\cdot)$ indicates the softmax function
and $b \in \mathbbm{R}^{l^{2}} $ is the bias term. 
\subsubsection{Objective function}
$\;$The proposed HAN is trained on the VQA-HAT dataset, which contains triplets of $(I,q,\alpha_{h})$, 
where $\alpha_{h}$ is the human attention for the given image-question pair $(I,q)$. 
 
In order to make the model-generated attention map get close to the human attention map, 
the Mean Squared Error (MSE) is used as the objective function. 
The MSE loss $L_{mse}$ for a given training triplet is defined as 
\begin{equation}
L_{mse}( I,q,a_{h}; \theta )=(\alpha_{h}'(I,q)-\alpha_{h})^{2},
\end{equation}
where $\theta$ represents all trainable parameters used in the HAN. 
The HAN is trained by minimizing the overall MSE loss on a given training set $\mathcal{D} = \{(I,q,\alpha_{h})\}$. 
\begin{equation}\label{eq:obj_mse}
\underset {\theta} {\mathrm{argmin}} \sum_{(I,q,a_{h})\in \mathcal{D}}   L_{mse}( I,q,\alpha_{h}; \theta ).
\end{equation}

\begin{figure}[tb!]
\centering
\noindent\includegraphics[width=1.0\columnwidth]{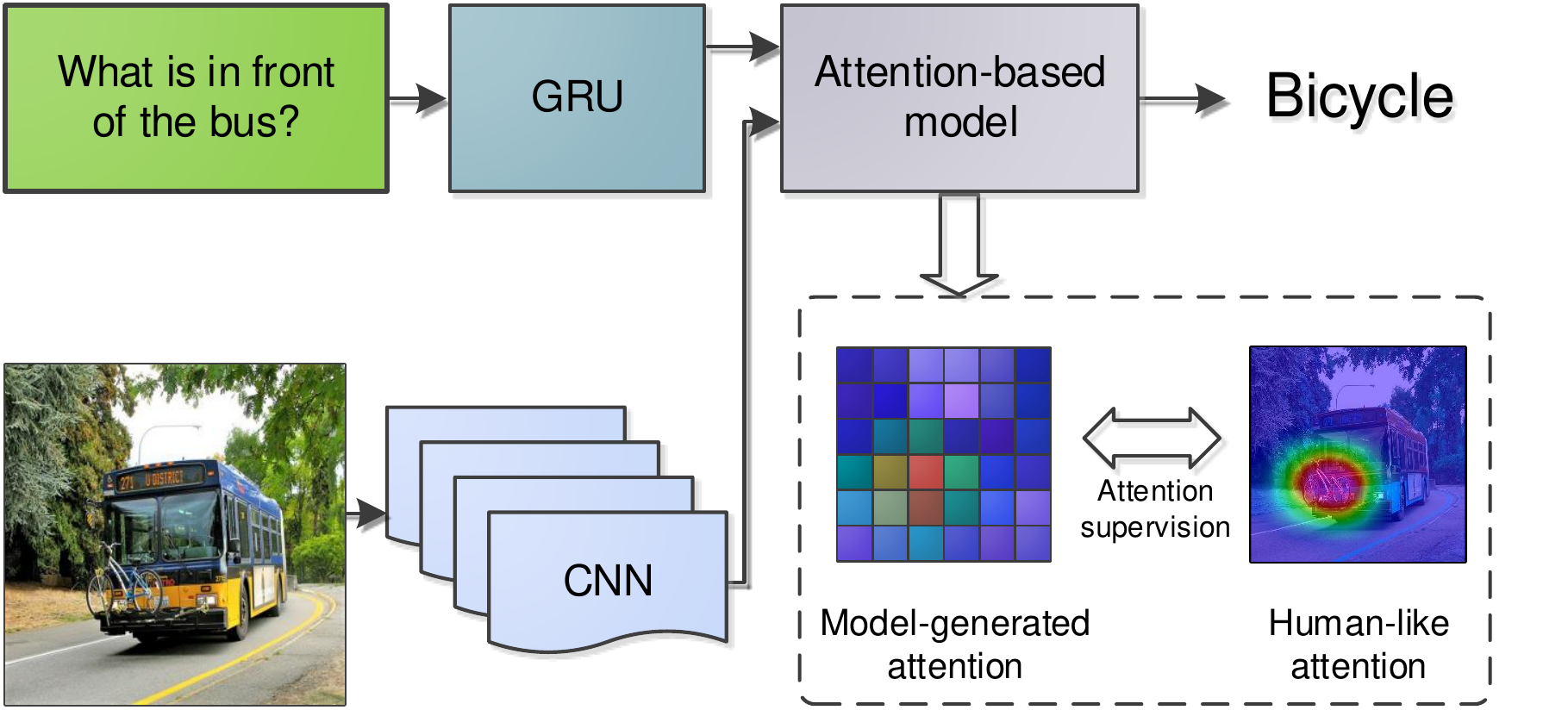}
\caption{Structure of the proposed supervised attention model. The model-generated attention explicitly learns from the human-like attention generated by the HAN.}
\label{fig:supervised_model}
\end{figure}

\subsection{Attention supervision for VQA}
In this section, the details of the VQA model are presented.
For comparison, an unsupervised attention model is discussed first. 
Afterwards, the proposed supervised attention model is introduced. 
\subsubsection{Unsupervised attention model}
The Multimodal Low-rank Bilinear attention networks (MLB) \cite{kim2016hadamard} is used as our baseline, 
due to its unsupervised attention mechanism and satisfactory performance. 
Given an image-question pair, the MLB first generates several attention maps for the given image-question pair and obtains the attentive visual feature. 
Then, the attentive visual feature and the question feature are fused by a low-rank bilinear pooling. 
The output is finally fed to a multi-way classifier to produce probabilities over the candidate answers.

Specifically, given an image feature matrix $F$ and a question feature vector $Q$, 
the attention maps are first computed as described in Eq. \ref{eq:att_fusion} and Eq. \ref{eq:att_generation}. 
Then, they are fed into a softmax layer to get the probability distribution over the image. 
Hence, a sequence of attention maps is obtained, 
denoted as $\alpha=\{\alpha_{1}, \alpha_{2},...,\alpha_{G}\}$. 
With the attention maps $\alpha$, 
the $s$-dimensional attentive visual feature is then calculated by 
\begin{equation}
\hat{F}=\bigparallel_{g=1}^G\sum_{i=1}^{l^2} softmax({\alpha}_{g}^i)F^i,
\end{equation}
where $\bigparallel$ denotes the concatenation operation of vectors, 
$\alpha_{g}^i$ indicates the $i$-th element of the $g$-th attention maps in $\alpha$, $F^i$ indicates the $i$-th row of an image feature map and the dimensionality of $\hat{F}$ is $s=m\times G$.

In order to predict the answer, 
the attentive visual feature vector $\hat{F}$ and the question vector $Q$ are combined through another low-rank bilinear pooling. 
The output of the low-rank bilinear pooling is fed to a linear layer followed by a softmax classifier to produce probabilities over the candidate answers. 
More formally, the answer probabilities are given by 
\begin{equation}
p(a\mid I, q;\Theta )=softmax(P(\sigma (U'Q)\circ \sigma (V'\hat{F}))),
\end{equation}
where $P\in \mathbbm{R}^{K\times r}$, $U'\in \mathbbm{R}^{r\times n}$ and $V'\in \mathbbm{R}^{r\times s}$ are trainable affine matrices.
The MLB model is trained by minimizing the cross-entropy loss over the training data. 
Given a training triplet of $(I,q,A)$, where $A$ is the ground truth of the predicted answer, 
the cross-entropy loss is defined as 
\begin{equation}
L_{cls}(I,q,A)=\frac{1}{K}\sum_{k=1}^{K}-log\;p(a_k\mid I,q\,;\Theta). 
\end{equation}
Here, $\Theta$ represents all parameters in the model and $A\in Ans$, 
where $Ans=\{a_1, a_2, ..., a_K\}$ is the set of candidate answers, $K$ is the size of the $Ans$. 
After training the model, the predicted answer of a given image-question pair $(I, q)$ is
\begin{equation}
\hat{a}=\mathrm{{arg max}}\;p(a\mid I, q;\Theta ).
\end{equation}
\subsubsection{Supervised attention model}
$\;$In the mentioned MLB model, attention maps are learned without any supervision, 
which may lead to inaccurate attention maps, as shown in Fig.\ref{fig:test_dev_result}. 
In order to remedy this problem, attention supervision is added to the unsupervised attention model. 
In this way, the attention maps that are generated during training can explicitly learn from the ground truth. 

Specifically, to train the human-like attention supervised model, the ground truth of the attention maps is needed. 
However, the human-labeled attention maps in the VQA-HAT dataset is not sufficient and manually annotating a new attention dataset is an expensive and time-consuming task. 
Therefore, the pre-trained HAN is used to generate attention maps for all the image-question pairs in the VQA v2.0 dataset. 
The generated attention map dataset for the VQA v2.0 dataset is named as Human-Like ATtention (HLAT) dataset. 
Fig.\ref{fig:han_examples} shows the comparison between the human attention and the human-like attention generated by the HAN. 
As it can be observed, the HAN-generated attention share similar attentive areas with the human attention, 
showing the promising results of using human-like attention as the ground truth for the attention learned in the models. 
Therefore, the HLAT is used for training the proposed supervised attention model. 
Fig.\ref{fig:supervised_model} shows the structure of the proposed supervised attention model. 

More formally, the triplet $(I,q,A)$ in the VQA v2.0 dataset is extended to $(I,q,A,\alpha_{h}')$, 
where $\alpha_{h}'$ is the generated human-like attention map corresponding to the image-question pair $(I,q)$. 
In order to employ attention supervision, 
a GRU is first employed to encode the sequence of attention maps $\alpha=\{\alpha_{1}, \alpha_{2},...,\alpha_{G}\}$ that is generated in the MLB to get a refined attention map. 
The output is then fed to a linear layer followed by a softmax layer to produce attention probabilities over the image.
The attention supervision loss is computed by
\begin{equation}
L_{att}(I,q,\alpha_{h}')=(softmax(W_h \cdot h_{GRU}(\alpha) + b_h ) - \alpha_{h}')^2,
\end{equation}
where $W_h \in \mathbbm{R}^{l^{2} \times p}$ is an affine transformation matrix, $b_h \in \mathbbm{R}^{l^{2}} $ is a bias term and $p$-dimensional $h_{GRU}(\alpha)$ indicates the last hidden state vector of the GRU, which is conceptually similar to the GRU defined in Eq.\ref{eq:att_gru}. 

The loss of the supervised attention model is the weighted sum of the classification loss and attention supervision loss. 
Given an extended training triplet of $(I,q,A,\alpha_{h}')$, the loss is defined as 
\begin{equation} \label{eq:total_loss}
L(I,q,A,\alpha_{h}')= L_{cls}(I,q,A) + \lambda \, L_{att}(I,q,\alpha_{h}'),
\end{equation}
where $\lambda$ is a trade-off parameter. The model is trained by minimizing the total loss of Eq.\ref{eq:total_loss}. 

\section{Experiment}
\subsection{Human Attention Network (HAN)}
\begin{figure*}[tb!]
\centering
\noindent\includegraphics[width=2\columnwidth]{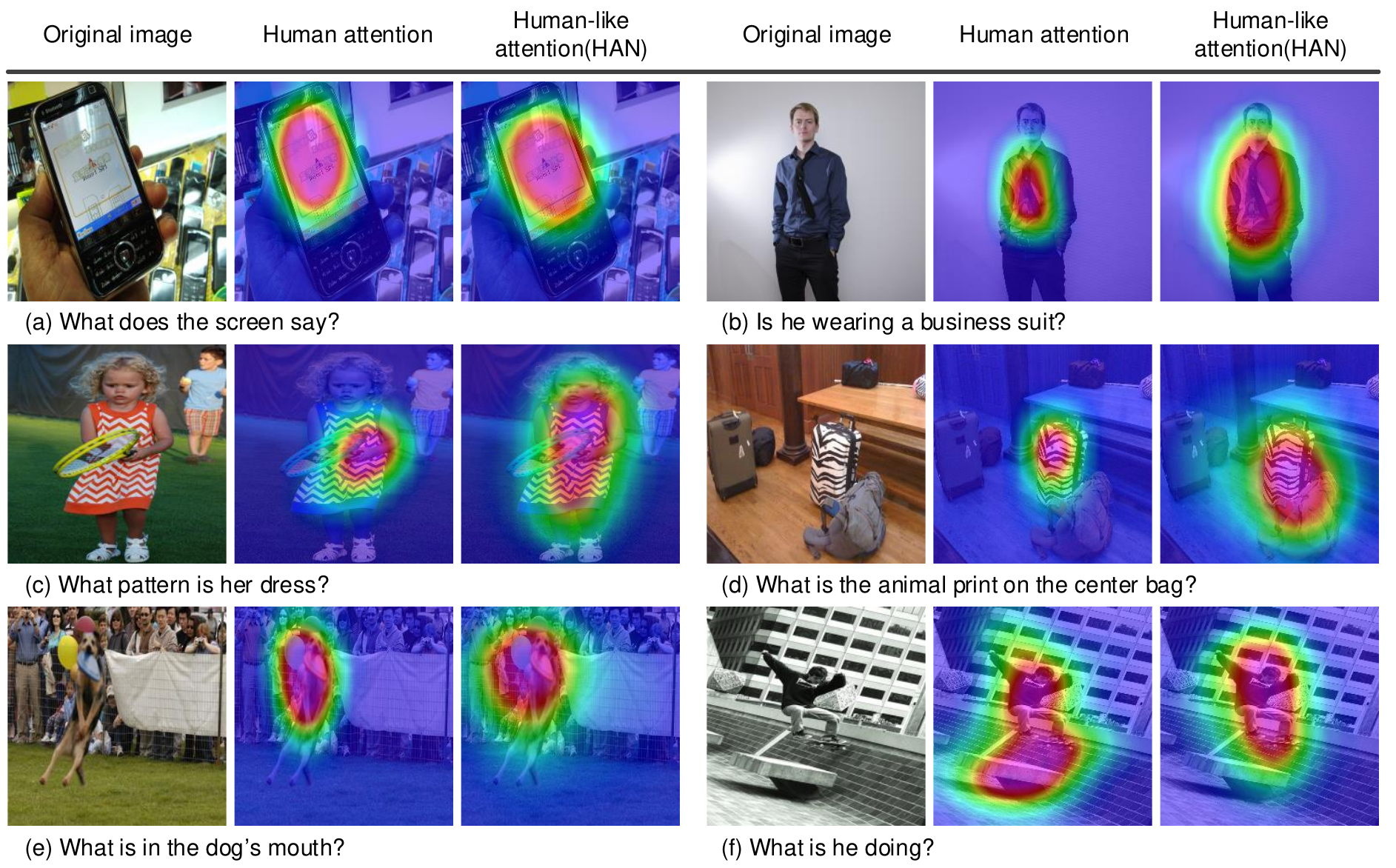}
\caption{Comparison between human attention and human-like attention generated by the HAN.}
\label{fig:han_examples}
\end{figure*}
\subsubsection{Dataset}
The HAN is evaluated on the recently released VQA-HAT dataset \cite{das2016human} as it is the only human-labeled attention dataset for VQA. 
The task of collecting the human attention data of the VQA-HAT dataset was implemented by 800 different annotators. 
The dataset contains human attention maps for around 60k image-question pairs in the VQA v1.0 dataset, which in total has approximately 600k image-question pairs in the dataset. 
Specifically, the annotators are shown a blurred image and a question about the image, 
and they need to sharpen the regions of the image that will help them to answer the question correctly. 
In the end, human attention maps were collected for 58,475 image-question pairs in the training set and 1,374 in the validation set of the VQA v1.0. 
The HAN is trained on the training set of the VQA-HAT and the experiment results are evaluated on the validation set of the VQA-HAT. 
Before training, 8 mislabeled attention maps whose value is zero were removed.
As the output of the HAN is processed through a softmax layer, 
all human attention maps are also fed to a softmax layer before training for modeling consistency. 
\subsubsection{Implementation details}
The experimental setup of \cite{kim2016hadamard} is also followed. 
The wording embedding matrix and the GRU for encoding words are initialized with the pre-trained skip-thought vector model \cite{kiros2015skip}. 
As a result, the question vectors have 2400 dimensions. 
The number of the glimpses in the HAN is set to 3, 
which shows the best performance. 
The size of the hidden state in the GRU for refining the attention maps is set to 512. 
The joint embedding size $d$ used for embedding images and questions is set to 1200. 
The Adam optimizer \cite{kingma2014adam} is used with a base learning rate of 3e-4.
The batch size is set to 64 and the iterations are set to 300k. 
Dropout is used with ratio 0.5. 
The network is implemented using the Torch7 \cite{torch}. 
\subsubsection{Evaluation metric: rank correlation}
The evaluation method provided in \cite{das2016human} is used. 
This evaluation method uses the mean rank-correlation coefficient to evaluate the quality of the attention maps, a higher value meaning a better quality. 
To be specific, first, both the machine-generated attention maps and the human labeled attention maps scaled to $14\times 14$. 
Then, the pixels are ranked according to their spatial attention value and finally, the correlation between the two ranked lists is computed. 
In \cite{das2016human}, in order to make the results more objective, 
three human attention maps are collected per image-question pair for the validation dataset
and the mean rank-correlation is computed among these three maps. 
Therefore, in this experiment, the reported correlation value is averaged over the three sets of attention maps that are labeled by different annotators.  
The rank-correlation in the experiment is calculated by 
\begin{equation}
\frac{1}{3}\sum_{i=1}^{3}(1-\frac{6\sum_{j=1}^{l^{2}} ({att'}^{j}-att_{i}^{j})^{2}}{l^{2}-l})
\end{equation}
where ${att'}^j$ is the $j$-th element of the predicted attention map and $att_{i}^{j}$ is the $j$-th element of the human attention map labeled by $i$-th annotator. 
$l$ denotes the size of the attention map, which is 14 in this experiment.

\begin{table} [tb!]
\renewcommand{\arraystretch}{1.2}
\caption{\textbf{Mean rank-correlation coefficients} A higher value means a better quality. Error bars show standard error of the mean. 
The results show that attention maps produced by the HAN have the highest correlation with human attention maps, surpassing the human performance by 0.045. 
}
\label{tab:Experiment-results-on}
\centering 
 \scalebox{0.9}{
\begin{tabular}{l r }
\toprule
\textbf{Model}& \textbf{Mean Rank-correlation}  \\
\cmidrule(l){1-2}
Random                                              & 0.000 $\pm$ 0.001\\
Human                                                & 0.623 $\pm$ 0.003 \\
SAN \cite{yang2016stacked}              & 0.249 $\pm$ 0.004 \\
HieCoAtt \cite{lu2016hierarchical}      & 0.264 $\pm$ 0.004  \\
HAN                               & \textbf{0.668 $\pm$ 0.001} \\
\bottomrule
\end{tabular}
 }
\end{table}
\begin{table} [tb!]
\renewcommand{\arraystretch}{1.2}
\caption{Results comparison between the HAN using the GRU and the HAN without using the GRU, where $G$ denotes the number of glimpses. As it can be observed, applying a GRU to fuse attention maps significantly improves the performance of the HAN.}
\label{tab:GRU_compare}
\centering 
 \scalebox{0.9}{
\begin{tabular}{l r }
\toprule
\textbf{Model}& \textbf{Mean Rank-correlation}  \\
\cmidrule(l){1-2}
HAN($G$=2,without GRU)             & 0.406 $\pm$ 0.001 \\
HAN($G$=2,with GRU)                  & \textbf{0.656 $\pm$ 0.002}  \\
HAN($G$=3,without GRU)             & 0.411 $\pm$ 0.001 \\
HAN($G$=3,with GRU)                  & \textbf{0.668} $\pm$ \textbf{0.001} \\
\bottomrule
\end{tabular}
 }
\end{table}
\subsubsection{Results}
Previous works \cite{kim2016hadamard,fukui2016multimodal} have explored the effect of the different number of glimpses in VQA. It was found that multiple glimpses may result in a better performance. Therefore, the HAN is also evaluated with different glimpses, namely 1, 2, 3 and 4, yielding a performance of 0.536, 0.656, 0.668 and 0.553 respectively. 
The number of glimpses is set to 3 in the experiment, as the model shows the best performance. 
The performance of the HAN is presented in Tab.\ref{tab:Experiment-results-on}, where the model achieves the best performance in predicting attention maps. 
As it can be observed, the performance of the HAN significantly outperforms the VQA models (SAN, HieCoAtt) by around 40$\%$, which indicates that the attention explicitly learned by the HAN is more human-like than the attention that is implicitly learned by traditional attention-based models. 
The correlation between human-labeled attention maps is 0.623, 
which shows that different annotators have different views about the relevant area in the image given the same question and the correctness of the attention maps is difficult to define. 
Therefore, for this model, making model-generated attention more human-like is the best choice. 
The experiment also shows that our generated attention has the highest correlation (0.668) with human attention, showing the ability of the HAN to generate human-like attention maps for image-question pairs. 

In order to verify the effectiveness of using the GRU to fuse attention maps, two experiments are conducted with the number of glimpses equal to 2 and 3. 
Specifically, in these two experiments, instead of fusing the attention maps using the GRU, the average attention map is obtained among several attention maps. 
The performance of our model with and without GRU is shown in Tab.\ref{tab:GRU_compare}. 
The HAN using the GRU significantly outperforms the HAN without using the GRU, showing the effectiveness of using the GRU to encode the attention maps. 
\subsection{Attention-based model for VQA}
\begin{figure*}[tb!]
\centering
\noindent\includegraphics[width=2\columnwidth]{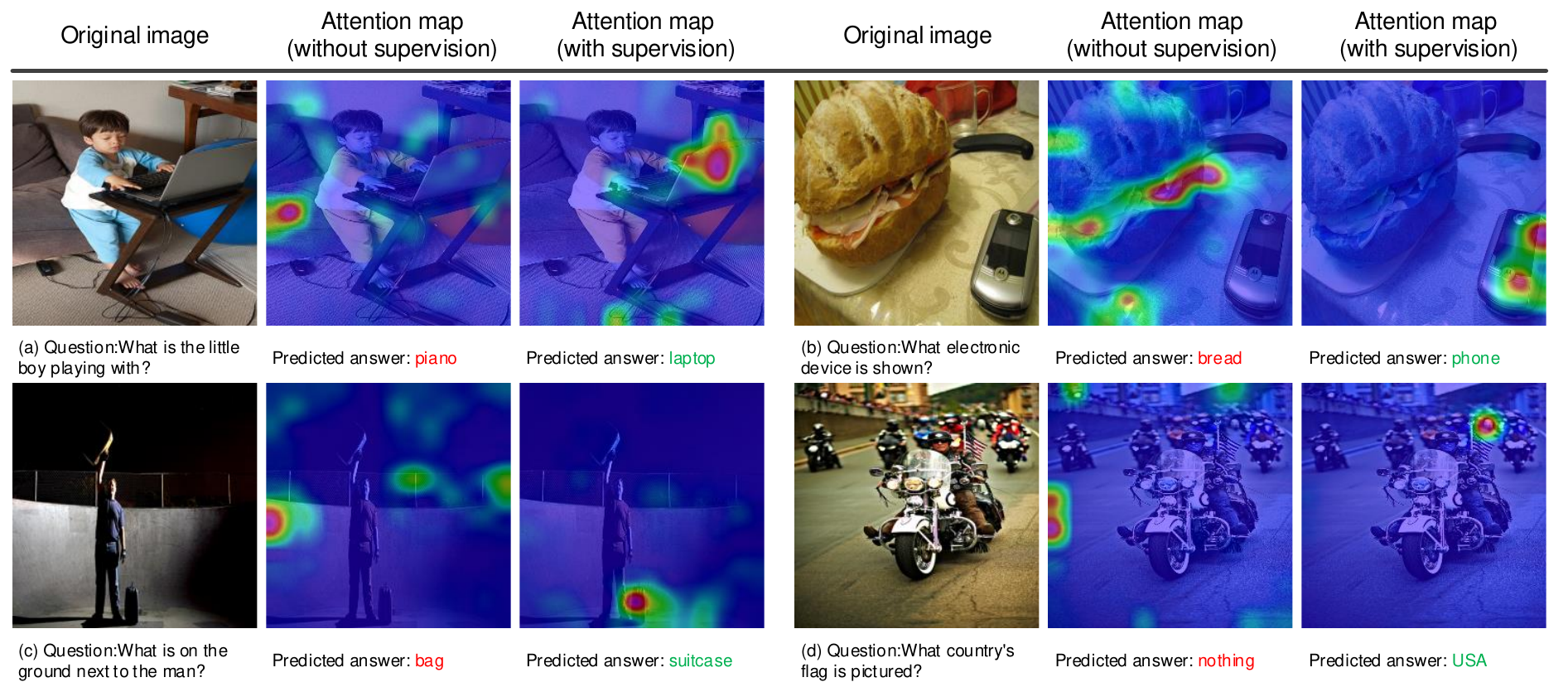}
\caption{Examples of generated attention maps and predicted answers. As shown in the figure, through attention supervision, the attention generated by attention-based model is more accurate, which results in better answers.}
\label{fig:test_dev_result}
\end{figure*}
\subsubsection{Dataset}
The unsupervised attention model and the supervised attention model are evaluated on the more recent VQA v2.0 dataset, 
which contains 443,757 image-question pairs in the training set,  214,354 in the validation set and 447,793 in the testing set.
This version of the dataset is more balanced in comparison to the VQA v1.0 dataset. 
Specifically, for every question, there are two similar images but they yield two different answers. 
Therefore, it forces models to focus more on visual information. 
The models are trained on the training and validation sets and tested on the test-dev set. 

When training the supervised attention model, the attention maps in the generated HLAT dataset are used as the ground truth of the attention maps. 
The HLAT contains attention maps for all the image-question pairs in the VQA v2.0 dataset. 
The third column of Fig.\ref{fig:han_examples} shows the examples of the attention maps in the HLAT dataset. 
\subsubsection{Implementation details}
The setup of the experiment is kept almost the same as above, excepting that the number of glimpses is set to 1 and 2, 
and the number of iterations is fixed to 300k in the VQA experiments. The top 3000 most frequent answers are used as possible outputs.
\begin{table} [tb!]
\renewcommand{\arraystretch}{1.2}
\caption{The VQA v2.0 test-dev results, where $G$ denotes the number of glimpses. }
\label{tab:VQA-experiment-results}
\centering 
\scalebox{0.82}{
\begin{tabular}{l r r r r}
\toprule
\textbf{Model}  & \textbf{Yes/No} & \textbf{Number} & \textbf{Other} & \textbf{Overall} \\
\cmidrule{1-5}
Prior     & 61.20  &  0.36  & 1.17 &  25.98     \\ 
Language-only                & 26.3    & 28.1   & 28.1  & 28.1\\
d-LSTM Q+norm I              & 26.3    & 28.1   & 41.85 & 54.22\\
\cite{antol2015vqa} &&&& \\
\midrule
unsupervised model($G$=1)             & 78.12    & 37.52   & 52.92  & 61.55\\
supervised model($G$=1)                 &78.03    & 37.93   & 53.14  & 61.66\\
unsupervised model($G$=2)             & 78.4   & 37.52   & 53.29  & 61.84\\
supervised model($G$=2)			    & \textbf{78.54}    & \textbf{37.94 }  & \textbf{53.38}  & \textbf{61.99}\\
\bottomrule
\end{tabular}
 }
\end{table}
\subsubsection{Evaluation metric}
The model results are evaluated using the provided accuracy metric \cite{antol2015vqa}. 
The accuracy of a predicted answer is given by 
\begin{equation}
\mathrm{min(\frac{\#humans\;that\;provided\ that\,answer}{3},1)}
\end{equation}
An answer is deemed 100$\%$ accurate if at least 3 annotators agree on the answer. 
Intuitively, this metric takes into account the consensus between the annotators. 
\subsubsection{Results}
Tab.\ref{tab:VQA-experiment-results} presents the accuracy comparison between the unsupervised model and the supervised model on the VQA v2.0 dataset. 
The accuracy of Prior, Language-only, deeper LSTM Q + norm I baseline models in \cite{antol2015vqa} are also presented for comparison.
As shown in this table, our supervised model outperforms the unsupervised model by 0.11$\%$ when the number of glimpses is set to 1 and by 0.15$\%$ when the number of glimpses is set to 2, showing the effectiveness of applying attention supervision to attention-based model. 
The supervised model achieves a much higher accuracy when the question is a counting problem, 
which indicates that the generated attention in the supervised model is more accurate and comprehensive. 
Fig.\ref{fig:test_dev_result} samples the attention maps and the answers that are predicted by both the unsupervised model and the supervised model. 
As it can be observed, after human-like attention supervision, the attention maps are more accurate and focused on the most relevant areas. 
It can therefore be concluded that more human-like attention helps to improve the performance of the model. 
\section{Conclusion}
In this work, we make a first attempt to verify the effectiveness of applying human-like attention supervision in VQA. 
The HAN is first proposed to generate human-like attention maps for image-question pairs. 
The HLAT dataset generated by the HAN can be potentially used as a supervision to help training attention-based models in VQA. 
The framework of the human-like attention supervision in the proposed supervised attention model can be easily applied to other attention-based models. 
The experiment results demonstrate that the supervised model outperforms the unsupervised model and through visualization, 
it is observed that the attention that is generated by the supervised attention model is more accurate, which yields better answers. 
In conclusion, explicit attention supervision indeed help attention-based models in VQA to get a better performance. 
\newpage 
\bibliographystyle{aaai}
\bibliography{egbib}
\newpage
\section{Supplementary Material}
\enlargethispage{-3.8cm}
\noindent\begin{picture}(0,0)
\put(0,-325){\begin{minipage}{\textwidth}
\centering
\includegraphics[width=0.9\linewidth]{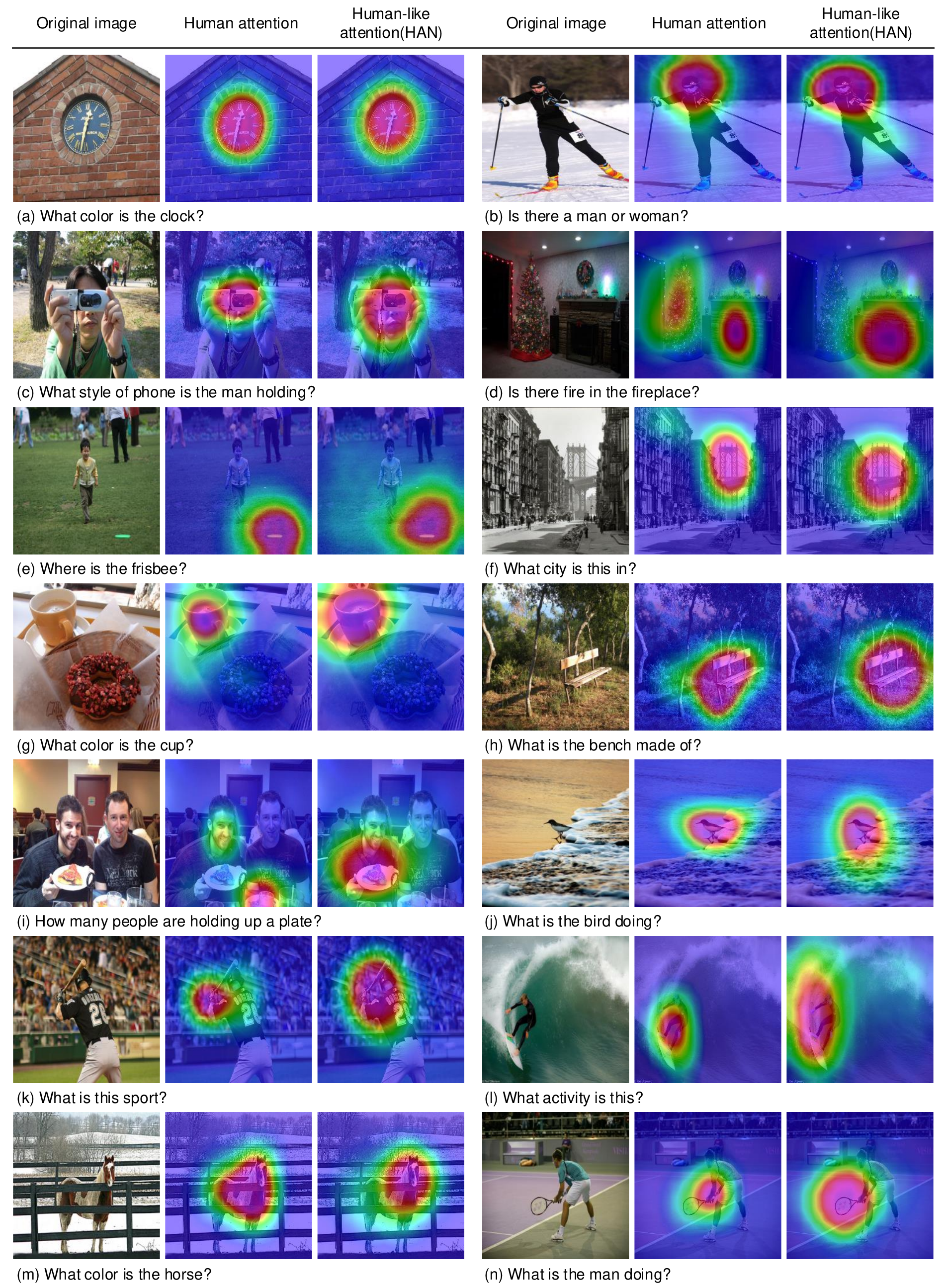}
\captionof{figure}{Examples of the human attention and the attention generated by the HAN. As it can be seen in the figure, the attention generated by the HAN has comparable quality as the human attention.}
\end{minipage}}
\end{picture}

\begin{figure*}[h]
\centering
\includegraphics[width=0.9\linewidth]{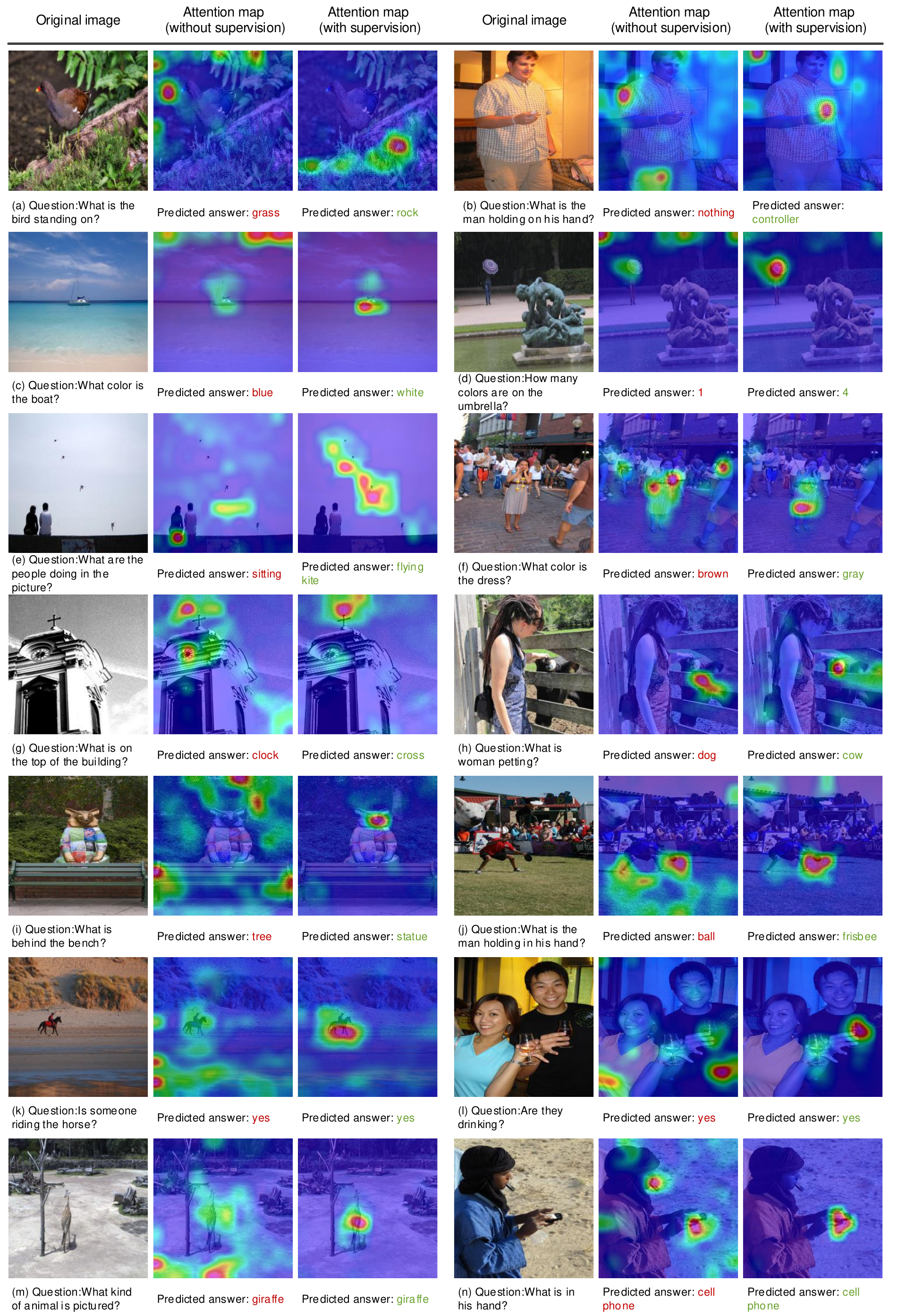}
\caption{Examples of the attention generated by the unsupervised model and the supervised model. As it can be observed, before attention supervision, the generated attention is relatively coarse and inaccurate. However, after attention supervision, the model generates more accurate attention and yields better answers. In (k), (l), (m) and (n), two models predict the same answers, however the supervised model has more confidence on the results as it has more accurate attention.}
\label{fig:image-to-text_sp}
\end{figure*}
\end{document}